\documentclass{article}

\usepackage[final]{ProbNum25} 

\usepackage[round]{natbib}

\usepackage{amsmath}
\usepackage{amssymb}
\usepackage{amsthm}
\usepackage{mathtools}
\usepackage{dsfont}
\newcommand{\1}{\mbox{1}\hspace{-0.25em}\mbox{l}}
\usepackage[nice]{nicefrac}
\newtheorem{theorem}{Theorem}[section]

\usepackage{todonotes}

\usepackage{paralist}
\usepackage{comment}
\usepackage{tcolorbox}

\usepackage[capitalise,nameinlink]{cleveref}
\usepackage{kantlipsum}

\probnumtitle{Fixing the Pitfalls of Probabilistic Time-Series Forecasting Evaluation by Kernel Quadrature}
\probnumauthors{%
\name{Masaki Adachi$^\star$}%
\affiliation{Lattice Lab, Toyota Motor Corporation / Machine Learning Research Group, University of Oxford}
\and%
\name{Masahiro Fujisawa$^\star$}%
\affiliation{The University of Osaka / Lattice Lab, Toyota Motor Corporation / RIKEN AIP}
\and%
\name{Michael A. Osborne}%
\affiliation{Machine Learning Research Group, University of Oxford}
}

\probnumabstract{
Despite the significance of probabilistic time-series forecasting models, their evaluation metrics often involve intractable integrations. The most widely used metric, the continuous ranked probability score (CRPS), is a strictly proper scoring function; however, its computation requires approximation. We found that popular CRPS estimators—specifically, the quantile-based estimator implemented in the widely used GluonTS library and the probability-weighted moment approximation—both exhibit inherent estimation biases. These biases lead to crude approximations, resulting in improper rankings of forecasting model performance when CRPS values are close. To address this issue, we introduce a kernel quadrature approach that leverages an unbiased CRPS estimator and employs cubature construction for scalable computation. Empirically, our approach consistently outperforms the two widely used CRPS estimators.
}

\begin{document}

\section{Introduction}
Time-series forecasting plays a central role in various applications, including finance~\citep{kim03, sezer2020financial}, healthcare~\citep{bui2018time, morid2023time}, and renewable energy~\citep{wang2019review, Dumas22, adachi2023bayesian}. Perfect prediction is inherently unattainable due to the difficulty of forecasting the future. Consequently, probabilistic modeling is often employed—not only to improve point prediction accuracy but also to capture the predictive distribution.

A wide range of probabilistic models has been proposed, including ARIMA~\citep{box1970}, Gaussian processes (GPs; \citet{Roberts13}), and deep learning-based models~\citep{Dumas22, Oskarsson24}. The key question is how to properly evaluate the accuracy of predictive distribution inference, rather than relying solely on point prediction metrics such as mean squared error. Since the observed value at time $t$ is an instantiation of an underlying random variable, the goal is to match the true generative distribution rather than overfitting to the observed test point.

Thus, researchers have long sought better metrics for probabilistic forecasting~\citep{matheson1976scoring, hersbach2000decomposition}, and there is now a consensus that the desired scoring rule should be \emph{strictly proper} \citep{gneiting2007strictly}. This condition ensures that the expected score is minimized when the predicted distribution matches the true generative distribution. Several strictly proper scoring rules exist, such as the Brier score \citep{Brier50}, but the Continuous Ranked Probability Score (CRPS; \citet{matheson1976scoring}) has gained particular popularity in the modern machine learning community~\citep{alexandrov2020gluonts, kollovieh2024predict, toth2024learning}. CRPS has a closed-form solution for commonly used parametric distributions, such as Gaussian and logistic distributions, making it especially suitable for evaluating GP models.

However, deep learning models typically do not rely on classical parametric distributions, necessitating the approximation of CRPS via sampling from the predictive distribution. The current sampling-based approach relies on a grid search over the quantile space, but we identify issues in its estimation bias. In particular, while we expect sample-based estimators to exhibit asymptotic convergence behavior, we demonstrate that a persistent bias exists between the true CRPS and its approximation—one that remains even with an infinite number of samples when the grid size is fixed.

To address this, we propose kernel quadrature as a principled approach to improving finite-sample estimators. We show that our method converges to the true CRPS faster than popular CRPS estimators while remaining free from estimation bias. Although any unbiased estimator can correct this, our kernel quadrature method further reduces the quadratic complexity to linear.

\section{Problem setting}
Let $\textbf{x}_{0:L} = (\textbf{x}_0, \cdots, \textbf{x}_L) \in \text{Seq}(\mathbb{R}^d)$ be an input time-series and $\textbf{y}_{0:L} = (\textbf{y}_0, \cdots, \textbf{y}_L) \in \text{Seq}(\mathbb{R})$ be a univariate output time series\footnote{We can extend the multivariate output, e.g., via multi-output GPs, but we describe only the univariate case.}. We assume a latent variable models $p_\theta(\textbf{y}_{0:L}) = \int p_\theta(\textbf{y}_{0:L}, \textbf{x}_{0:L}) \text{d} \textbf{x}$, where $\textbf{y}_{0:L} \sim p(\textbf{y}_{0:L})$ represents the true underlying distribution. We define the training dataset as $\textbf{D}_{0:L} = (\textbf{x}_{0:L}, \textbf{y}_{0:L})$ and the test dataset as $\textbf{D}_{L+1:L+T} = (\textbf{x}_{L+1:L+T}, \textbf{y}_{L+1:L+T})$, where $L$ and $T$ denote the training and test sizes, respectively.

We consider a set of candidate autoregressive models\footnote{The model is conditioned on the previous state; see \citet{toth2024learning} for details.}, $f^{(i)}(x) = p_{\theta}(y \mid x, \textbf{D}_{0:L})$, where different models $f^{(i)}$ and $f^{(j)}$ are not merely distinguished by parameterization $\theta$ but belong to entirely different function classes (e.g., Gaussian processes and diffusion models). Our goal is to identify the most plausible model given a collection of datasets $\{ \textbf{D}^{(k)}_{0:L_k+T_k} \}_{k=1}^K$. 

At test time, for each dataset $\textbf{D}^{(k)}$, we have access only to the test input sequence $\textbf{x}_{L_k+1:L_k+T_k}$, where the dataset size $\lvert \textbf{D}^{(k)} \rvert = L_k + T_k$ varies across datasets. 
Given a probabilistic prediction $f^{(i)}(x_l)$ and the hidden ground truth $\textbf{y}_l$ at the $l$-th timestep, we evaluate the performance of the $i$-th model using the CRPS metric:
\begin{align}
    \text{CRPS}(F, y_l) := \int_{-\infty}^\infty 
    \left( F(y^\prime) - \1_{y^\prime < y}\right)^2
    \text{d} y^\prime, \label{eq:crps}
\end{align}
where $F(y)$ is the cumulative density function (CDF) of the probability distribution of $y$, and $\1_{y^\prime < y}$ is an indicator function that returns 1 if the condition $y^\prime < y$ is satisfied and 0 otherwise.

It is common to use the sample average of CRPS as the scoring rule: $S(f^{(i)} \mid \textbf{D}^{(k)}, r_j) = \nicefrac{1}{T_k - L_k}\sum_{l=L_k}^{T_k} $ $\text{CRPS}(F^{(k)}, y_l)$, where $r_j \in \mathcal{R}$ is the $j$-th random seed for computation, sampled uniformly from $\mathcal{U}(\mathcal{R})$. To assess the effect of random seeds, we further iterate this scoring procedure by re-training the model $f^{(i)}$ with different random seeds $r_j$. The final evaluation metrics reported in the paper are the sample mean: $\overline{S} = \mathbb{E}_{r_j \sim \mathcal{U}(\mathcal{R})} [S(f^{(i)} \mid \textbf{D}^{(k)}, r_i)]$ and the variance:\\ $\mathbb{V}_{r_j \sim \mathcal{U}(\mathcal{R})}[S(f^{(i)} \mid \textbf{D}^{(k)}, r_i)]$. The performance ranking of the proposed model $f^{(i)}$ is primarily based on the expectation $\overline{S}$. 
As such, the goal of this task is to minimize the following integral approximation error: 
\small
\begin{align*}
    \Bigg| &\mathbb{E}_{r_j \sim \mathcal{U}(\mathcal{R})}[S(f^{(i)} \mid \textbf{D}^{(k)}, r_i)] - \mathbb{E}_{r_i \sim \mathcal{U}(\mathcal{R})}[\hat{S}(f^{(i)} \mid \textbf{D}^{(k)}, r_i)] \Bigg|,
\end{align*}
\normalsize
where $\hat{S}$ denotes the approximated sample-mean CRPS\footnote{GP has a closed-form CRPS, making this error zero. However, for a fair comparison with deep learning models, it is common to evaluate GP using an approximated CRPS obtained via sampling.}.

\subsection{Known results and approximation}
\begin{figure}[t!]
    \centering
    \includegraphics[width=0.6\hsize]{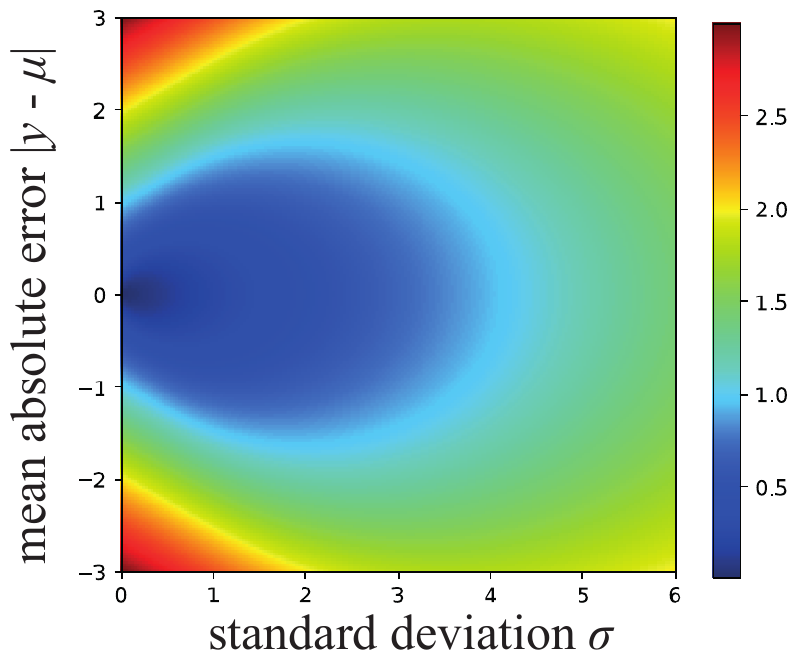}
    \caption{The exact CRPS for given Gaussian parameters.
    }
    \label{fig:convex}
\end{figure}
\paragraph{Exact CRPS with Gaussian}
CRPS has a closed-form solution for a univariate Gaussian predictive distribution, where $y_l \sim \mathcal{N}(m_l, \sigma^2_l)$ at the $l$-th timestep (see Eq.~(5) in \citet{gneiting2005calibrated}). Let $z_l := \nicefrac{y_l - m_l}{\sigma_l}$ be the standardized output, then we have:
\begin{align}
    \text{CRPS}(F, y_l) = \sigma_l \left[ 
    z_l(2 \Phi(z_l) - 1) + 2 \phi(z_l) - \frac{1}{\sqrt{\pi}}
    \right], \label{eq:exact}
\end{align}
where $\Phi(\cdot)$ and $\phi(\cdot)$ denote the CDF and probability density function (PDF) of the standard normal distribution $\mathcal{N}(0,1)$, respectively. This closed-form expression offers us to directly compute the exact CRPS for GP models. Similarly, closed-form solutions are available for popular parametric distributions (see Appendix B in \citet{taillardat2016calibrated} for a complete list)\footnote{The closed-form CDF is limited to univariate distributions. Thus, multivariate time series require approximation.}.

Eq.\eqref{eq:exact} provides an intuitive understanding of CRPS. Since the variance $\sigma_l$ is a multiplicative factor in all terms, a smaller predictive variance leads to a lower CRPS. Additionally, the expression inside the brackets is convex with respect to the standardized output $z_l$, attaining its minimum at $z_l = 0$. This implies that a lower mean absolute error, $\lvert y_l - m_l \rvert$, results in a better score. Fig.\ref{fig:convex} illustrates this intuition. Thus, CRPS serves as a reasonable scoring rule for evaluating both predictive accuracy and the tightness of predictive variance.

\paragraph{Approximating CRPS with quantile loss.}
Exact computation of CRPS is not always feasible for all predictive models. In particular, deep learning-based probabilistic forecasting models, such as diffusion models, do not have closed-form predictive distributions. Thus, CRPS must be estimated from i.i.d. function samples.

There are two common approaches for sample-based CRPS approximation. The most widely used method is the quantile loss reformulation \citep{kollovieh2024predict}:
\begin{align}
    \text{CRPS}(F, y_l) = \int_{0}^1 2 \Lambda_\kappa (F^{-1}(\kappa), y_l) \text{d} \kappa, \label{eq:quantile}
\end{align}
where $F^{-1}$ is the quantile function (also known as the inverse CDF), and $\Lambda_\kappa(q, y) = (\kappa - \1_{y < q})(y - q)$ represents the pinball loss for a given quantile level $\kappa$. To approximate the quantile function, we typically use the empirical CDF~\citep{dekking2006modern}.

The estimation procedure consists of two steps: First, we draw $M$ i.i.d. function samples at the test input time series,
$\textbf{f}_l = \{f^{(i)}_{m}(x_l)\}_{m=1}^M$, and then estimate the empirical CDF, $\hat{F}(y) = \nicefrac{1}{M} \sum_{m=1}^M \1_{f^{(i)}_{m}(x_l) \leq y}$. 
Next, we discretize the quantile levels $\kappa_\ell \in \mathcal{K}$ using a finite set, $\mathcal{K} = (\kappa_1,..., \kappa_Q) = (\nicefrac{1}{2Q},...,\nicefrac{2Q-1}{2Q})$. 
Using this discretization, we approximate the CRPS in Eq.~\eqref{eq:crps} as:
\begin{align}
    \widehat{\text{CRPS}}(\hat{F}, y_l) = \frac{1}{Q}\sum_{\kappa_\ell \in \mathcal{K}} 2 \Lambda_{\kappa_\ell}(\hat{F}^{-1}(\kappa_\ell), y_l). \label{eq:empirical_crps}
\end{align}
Due to the high computational cost of deep learning models, the sample sizes for all approximation steps are limited~\citep{kollovieh2024predict, toth2024learning}. For the empirical CDF, the number of function samples is typically set to $M=100$, and the quantile levels are uniformly discretized into nine points, 
$\mathcal{K} = (0.1,0.2,0.3,0.4,0.5,0.6,0.7,0.8,0.9)$. Also, due to the computational burden of retraining deep learning models, the number of random seeds is limited to $\lvert \mathcal{R} \rvert = 3$\footnote{We note that the quantile prediction approach  \citep{cai2002regression} can accelerate quantile-based CRPS evaluation.}.

\paragraph{Approximating the CRPS with the PWM.}
\begin{figure}[t!]
    \centering
    \includegraphics[width=0.8\hsize]{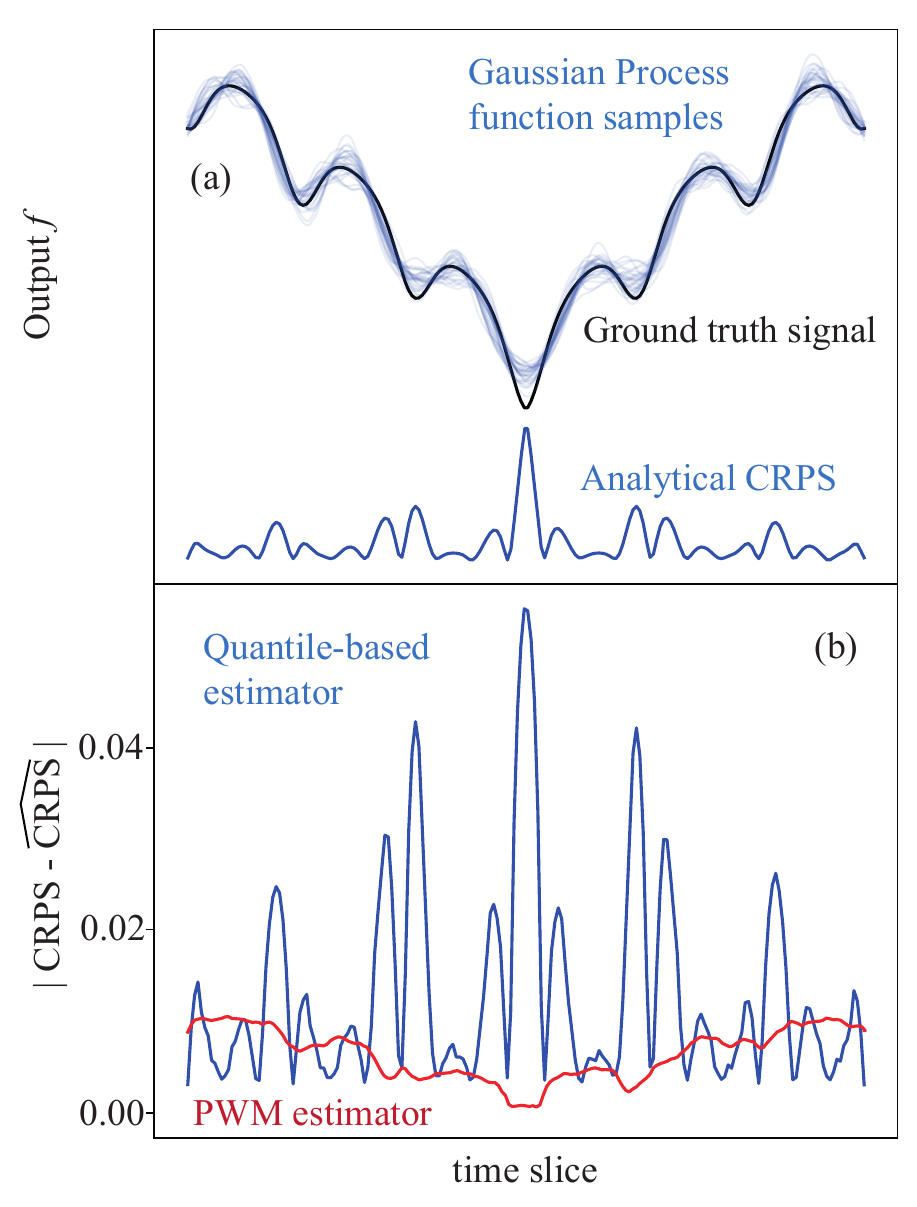}
    \caption{Illustrative example: (a) Ackley function fitted by a GP with its function samples and the analytical CRPS computed using Eq.\eqref{eq:exact}, (b) slicewise CRPS estimation error for the quantile-based estimator (Eq.\eqref{eq:quantile}) and the probability-weighted moment (PWM) estimator (Eq.~\eqref{eq:pwm}).
    }
    \label{fig:example}
\end{figure}
Another widely used approximation is the probability-\\
weighted moment (PWM; ~\citet{taillardat2016calibrated}):
\small
\begin{equation}
\begin{aligned}
    &\text{CRPS}(F, y_l)\\
    =& \underbrace{\mathbb{E}_{y \sim \mathbb{P}(f \mid x_l)}[\lvert y - y_l \rvert]}_\text{error term} + \underbrace{\mathbb{E}_{y \sim \mathbb{P}(f \mid x_l)}[y]}_\text{mean term} - \underbrace{2\mathbb{E}_{y \sim \mathbb{P}(f \mid x_l)}[y F(y)]}_\text{CDF term}. \label{eq:pwm}
\end{aligned}
\end{equation}
\normalsize
The advantage of this approach is that it simplifies CRPS estimation into a straightforward Monte Carlo (MC) integration. For a Gaussian predictive distribution, each term has a closed-form expression:
\begin{equation}
\begin{aligned}
    \mathbb{E}_{y \sim \mathbb{P}(f \mid x_l)}[\lvert y - y_l \rvert] 
    &= 
    \sigma_l [z_l (2 \Phi(z_l) - 1) + 2 \phi(z_l)],\\
    \mathbb{E}_{y \sim \mathbb{P}(f \mid x_l)}[y] 
    &=
    \mu_l,\\
    \mathbb{E}_{y \sim \mathbb{P}(f \mid x_l)}[y F(y)]
    &=
    \frac{1}{2} \left(\mu_l + \frac{\sigma_l}{\sqrt{\pi}}\right). 
\end{aligned}\label{eq:pwm_error}
\end{equation}
As such, this approximation equals to Eq.~(\ref{eq:exact}).

\section{Pitfalls of CRPS approximation}\label{sec:pitfall}
\begin{figure}[t!]
    \centering
    \includegraphics[width=0.8\hsize]{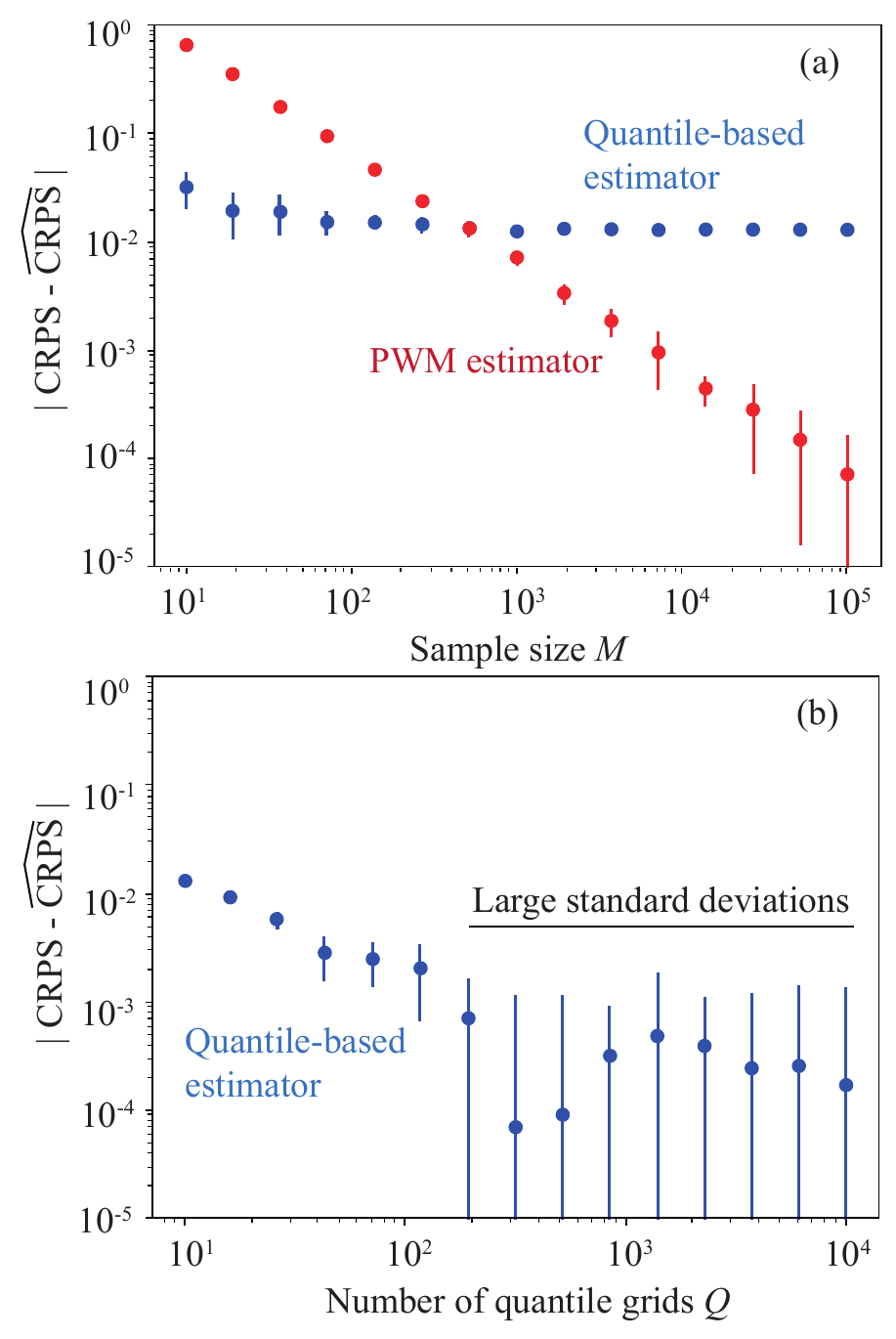}
    \caption{Convergence rate analysis (mean $\pm$ 1 standard deviation over 10 random seeds). (a) While the PWM estimator converges with respect to the sample size $M$, the quantile-based estimator does not for a fixed $Q=9$. (b) The quantile-based estimator exhibits convergence with respect to the number of quantile grids $Q$, but it plateaus around $Q = 100$ for $M=10^2$.
    }
    \label{fig:convergence}
\end{figure}
\subsection{Quantile or PWM?}
Given the two approximation methods, a natural question arises: which one should we use for evaluating time-series models? Due to the popularity of the GluonTS library~\citep{alexandrov2020gluonts}, the quantile-based estimator has dominated recent publications. However, we argue that this approach falls into hidden pitfalls. To illustrate these pitfalls, we first analyze a toy example to understand the typical behavior of CRPS approximation and identify the sources of evaluation bias.

Fig.~\ref{fig:example} explains the set up: we use the Ackley function as the test function~\citep{ackley2012connectionist} and a GP time-series model as the forecasting model. For simplicity, we randomly sample nine points from the domain and fit a GP model to these data points. We then draw $M$ function samples over $T=200$ test points. Since the GP predictive distribution is Gaussian, we compute the analytical CRPS using Eq.~(\ref{eq:exact}) (see Fig.~\ref{fig:example}(a)). Next, we compare the two approximation methods—quantile-based and PWM estimators—based on $M$ samples. Fig.~\ref{fig:example}(b) shows the estimation error across the domain. Notably, the error of the quantile-based estimator closely follows the shape of the analytical CRPS, whereas the error of the PWM estimator roughly follows the predictive mean of the GP model. Ideally, an unbiased estimator should exhibit no systematic pattern over time. This result suggests that both the quantile and PWM estimators introduce estimation bias.

The bias issue becomes more evident when we examine the convergence rate with respect to the sample size $M$ in Fig.~\ref{fig:convergence}(a). While the PWM estimator exhibits asymptotic convergence, the quantile-based estimator plateaus, indicating clear estimation bias. Interestingly, in the small sample size regime ($M < 10^3$), the quantile-based estimator shows lower errors, but this advantage disappears at larger sample sizes.

This leads to the first pitfall: \emph{spurious supremacy of quantile-based estimator}. This phenomenon misguides the community into adopting the current de facto standard setting (quantile estimator with $M=100$ and $Q=9$). Under this setting, the quantile method appears superior because its error is lower. Additionally, its stability under varying sample sizes reinforces the misconception that $M=100$ is sufficiently large and that further sampling is unnecessary.

However, this plateau occurs because the default quantile grid size, $Q = 9$, is too coarse. Fig.~\ref{fig:convergence} (b) confirms that increasing the grid size $Q$ leads to asymptotic convergence. Hence, the quantile estimator requires a balance between grid size $Q$ and sample size $M$.

From a computational perspective, the quantile estimator has complexity $\mathcal{O}(Q M \log M)$, whereas the PWM estimator has a lower complexity of $\mathcal{O}(M \log M)$. Thus, achieving convergence with the quantile estimator is more computationally expensive. Ideally, we should prefer the simpler PWM estimator.

\subsection{Why is PWM worse in small samples?}
\begin{figure}[t!]
    \centering
    \includegraphics[width=0.8\hsize]{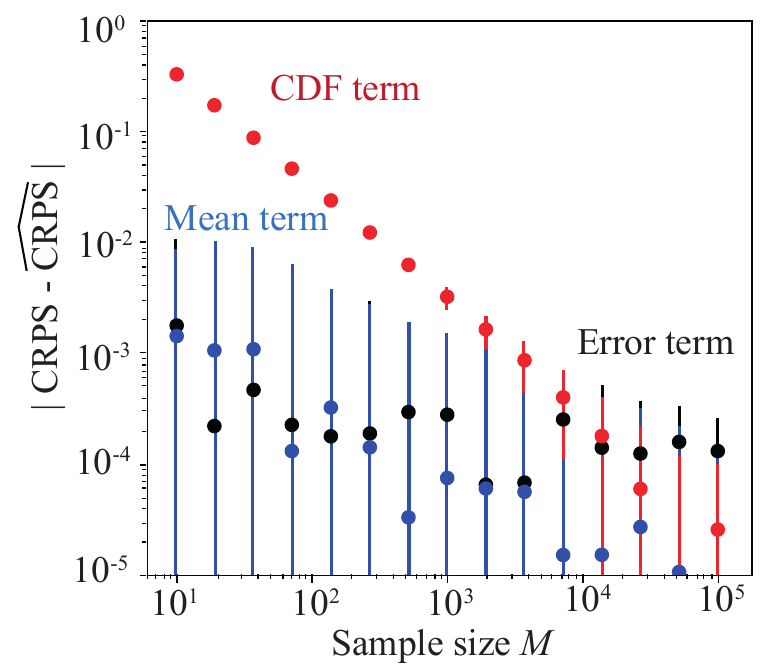}
    \caption{The decomposition of PWM estimation errors. CDF term dominates other two terms in the small sample regime.
    }
    \label{fig:err_analysis}
\end{figure}
\begin{table*}
    \centering
    \caption{(Copied from \cite{toth2024learning} Table 1): Forecasting results on eight benchmark datasets ranked by CRPS. The best and second best models have been shown as \textbf{bold} and \underline{underlined}, respectively.}
    \resizebox{1.0\textwidth}{!}{
    \begin{tabular}{lcccccccc}
        \toprule
         method&  Solar & Electricity & Traffic & Exchange & M4 & UberTLC & KDDCup & Wikipedia\\
         \midrule
         Seasonal Naïve & 0.512 $\pm$ 0.000 & 0.069 $\pm$ 0.000 & 0.221 $\pm$ 0.000 & 0.011 $\pm$ 0.000 & 0.048 $\pm$ 0.000 & 0.299 $\pm$ 0.000 & 0.561 $\pm$ 0.000 & 0.410 $\pm$ 0.000\\
         ARIMA & 0.545 $\pm$ 0.006 & - & - & \textbf{0.008 $\pm$ 0.000} & 0.044 $\pm$ 0.001 & 0.284 $\pm$ 0.001 & 0.547 $\pm$ 0.004 & - \\
         ETS & 0.611 $\pm$ 0.040 & 0.072 $\pm$ 0.004 & 0.433 $\pm$ 0.050 &  \textbf{0.008 $\pm$ 0.000} & 0.042 $\pm$ 0.001 & 0.422 $\pm$ 0.001 & 0.753 $\pm$ 0.008 & 0.715 $\pm$ 0.002 \\
         Linear & 0.569 $\pm$ 0.021 & 0.088 $\pm$ 0.008 & 0.179 $\pm$ 0.003 & 0.011 $\pm$ 0.001 & 0.039 $\pm$  0.001 & 0.360 $\pm$ 0.023 & 0.513 $\pm$ 0.011 & 1.624 $\pm$ 1.114\\
         \midrule
         DeepAR & 0.389 $\pm$ 0.001 & \underline{0.054 $\pm$ 0.000} & \underline{0.099 $\pm$ 0.001} & 0.011 $\pm$ 0.003 & 0.052 $\pm$ 0.006 & \textbf{0.161} $\pm$ \textbf{0.002} & 0.414 $\pm$ 0.027 & 0.231 $\pm$ 0.008 \\
         MQ-CNN & 0.790 $\pm$ 0.063 & 0.067 $\pm$ 0.001 & - & 0.019 $\pm$ 0.006 & 0.046 $\pm$ 0.003 & 0.436 $\pm$ 0.020 & 0.516 $\pm$ 0.012 & 0.220 $\pm$ 0.001 \\
         DeepState & 0.379 $\pm$ 0.002 & 0.075 $\pm$ 0.004 & 0.146 $\pm$ 0.018 & 0.011 $\pm$ 0.001 & 0.041 $\pm$ 0.002 & 0.288 $\pm$ 0.087 & - & 0.318 $\pm$ 0.019 \\
         Transformer & 0.419 $\pm$ 0.008 & 0.076 $\pm$ 0.018 & 0.102 $\pm$ 0.002 & \underline{0.010 $\pm$ 0.000} & 0.040 $\pm$ 0.014 & 0.192 $\pm$ 0.004 & 0.411 $\pm$ 0.021 & \textbf{0.214} $\pm$ \textbf{0.001} \\
         TSDiff & \underline{0.358 $\pm$ 0.020} & \textbf{0.050} $\pm$ \textbf{0.002} & \textbf{0.094} $\pm$ \textbf{0.003} & 0.013 $\pm$ 0.002 &  0.039 $\pm$ 0.006 & \underline{0.172 $\pm$ 0.008} & 0.754 $\pm$ 0.007 & \underline{0.218 $\pm$ 0.010} \\
         \midrule
         SVGP & \textbf{0.341 $\pm$ 0.001} & 0.104 $\pm$ 0.037 & - & 0.011 $\pm$ 0.001 & 0.048 $\pm$ 0.001 & 0.326 $\pm$ 0.043 & 0.323 $\pm$ 0.007 & - \\
         DKLGP & 0.780 $\pm$ 0.269 & 0.207 $\pm$ 0.128 & - & 0.014 $\pm$ 0.004 & 0.047 $\pm$ 0.004 & 0.279 $\pm$ 0.068 & 0.318 $\pm$ 0.010 & - \\
         \midrule
         RS$^3$GP & 0.377 $\pm$ 0.004 & $0.057 \pm 0.001$ & $0.165 \pm 0.001$ & $0.012 \pm 0.001$ & \underline{0.038 $\pm$ 0.003} & $0.354 \pm 0.016$ & \underline{0.297 $\pm$ 0.007} & $0.310 \pm 0.012$ \\
         VRS$^3$GP & $0.366 \pm 0.003$ & 0.056 $\pm$ 0.001 & $0.160 \pm 0.002$ & 0.011 $\pm$ 0.001 & \textbf{0.035} $\pm$ \textbf{0.001} & $0.347 \pm 0.009$ & \textbf{0.291} $\pm$ \textbf{0.015} & $0.295 \pm 0.005$ \\
         \bottomrule
    \end{tabular}
    }
    \label{tab:crps}
\end{table*}
The key question is why the PWM estimator performs worse in the small sample size regime. The empirical convergence rate in Fig.~\ref{fig:convergence}(a) is approximately $\mathcal{O}(\nicefrac{1}{M})$, which is faster than the well-known Monte Carlo (MC) integration rate of $\mathcal{O}(\nicefrac{1}{\sqrt{M}})$. This is unexpected because the second term of the PWM estimator in Eq.~(\ref{eq:pwm}) is a pure MC integral, implying a convergence rate of $\mathcal{O}(\nicefrac{1}{\sqrt{M}})$. Consequently, the overall convergence rate of PWM should be limited by this slowest component.

A possible explanation is that a certain bottleneck term in Eq.(\ref{eq:pwm}), which has a better convergence rate but a large constant, slows down the overall convergence. Fig.\ref{fig:err_analysis} decomposes the convergence rate of each error term using the closed-form expressions in Eq.~(\ref{eq:pwm_error}). The analysis reveals that the CDF term dominates the error—it follows a faster $\mathcal{O}(1/M)$ rate but with a large constant—while the other two terms approximately follow the slower $\mathcal{O}(\nicefrac{1}{\sqrt{M}})$ rate but with a smaller constant.

This leads to the second pitfall: \emph{the hidden bottleneck of the PWM estimator}. Unlike the first pitfall, this issue is more subtle and requires a step-by-step examination.
First, consider the CDF term in Eq.~(\ref{eq:pwm}). It is a nonlinear functional of the estimated CDF $\hat{F}$, making it a type of plug-in estimator. As previously discussed, we typically approximate the CDF using the empirical CDF, which has a well-known asymptotic convergence rate for i.i.d. samples:
\begin{align}
    \hat{F}(y) - \epsilon \leq F(y) \leq \hat{F}(y) + \epsilon, \,\,\, \text{where} \,\,\, \epsilon = \sqrt{\frac{\ln \nicefrac{2}{\alpha}}{2 M}},
\end{align}
with at least probability $1 - \alpha$. This bound is well known as the Dvoretzky–Kiefer–Wolfowitz (DKW) inequality~\citep{dvoretzky1956asymptotic}, which states that the convergence rate of the empirical CDF is also $\mathcal{O}(\nicefrac{1}{\sqrt{M}})$. However, this contradicts our observation in Fig.~\ref{fig:err_analysis}.

This discrepancy can be explained by plug-in bias, a common issue where the convergence rate of a plug-in estimator for a nonlinear functional introduces an asymptotic bias term that is independent of the finite-sample estimation error. Even though the empirical CDF weakly converges to the true CDF, the nonlinearity in the CRPS functional causes the expectation of the plug-in estimator to deviate from the true CRPS by a constant bias term. Therefore, to fully leverage the PWM estimator, we need to correct for this bias.

\subsection{The source of plug-in bias}
Recall that the empirical CDF is defined as: $\hat{F}(y_i) = \frac{1}{M} \sum_{j=1}^M \1_{y_j \leq y_i}$. For simplicity, we denote the CDF term in Eq.(\ref{eq:pwm}) as $\mathbb{E}[C(\hat{F})]$ and the corresponding analytical solution as $C(F)$ from Eq.(\ref{eq:pwm_error}). We have:
\begin{align*}
    \mathbb{E}[C(\hat{F})] 
    &= \frac{1}{M} \sum_{i=1}^M y_i \hat{F}(y_i),\\
    &= \frac{1}{M^2} \sum_{i=1}^M \sum_{j=1}^M y_i \1_{y_j \leq y_i},\\
    &= \frac{1}{M^2} \sum_{i, j = [M]} h(y_i, y_j),\\
    &= \frac{1}{M^2} \left( \underbrace{M \mathbb{E}[h(y_i,y_i)]}_\text{diagonal} + \underbrace{M(M-1) \mathbb{E}[h(y_i,y_i)]}_\text{off-diagonal} \right),
\end{align*}
where $h(y_i, y_j) = y_i \1_{y_j \leq y_i}$.
We then have:
\begin{align*}
    &\mathbb{E}[h(y_i,y_i)] = \mathbb{E}[y \1_{y \leq y}] = \mathbb{E}[y] = \mu_l,\\
    &\mathbb{E}[h(y_i,y_i)] = \mathbb{E}[y_i \1_{y_i \geq y_j}] = C(F).
\end{align*}
Thus, we observe that the off-diagonal term corresponds to the true value $C(F)$, yet the estimator $C(\hat{F})$ unnecessarily includes an additional diagonal term. Using these identities, we obtain:
\begin{align*}
    \text{plug-in bias} := \mathbb{E}[C(\hat{F})] - C(F) = \frac{1}{M}\left( \mu_l - C(F) \right).
\end{align*}
This explains our observations. In Fig.~\ref{fig:example}(b), we see a $\mu_l$-dependent bias term, while in Fig.~\ref{fig:err_analysis}, the CDF term exhibits an $\mathcal{O}(1/M)$ convergence rate. These artifacts arise solely due to the inclusion of the diagonal term.

\subsection{Why does this error matter?}
These errors are significant because they are on the same order of magnitude as the differences between forecasting models. As an example, we reference the experimental results from Table 1 in \citet{toth2024learning}, which uses CRPS as implemented in the GluonTS library (quantile-based estimator with $M=100$ and $Q=9$). The reported differences between models are roughly in the range of $10^{-1}$ to $10^{-3}$, while the CRPS approximation errors at the default setting are on the order of $10^{-1}$ to $10^{-2}$. In other words, crude CRPS approximations can lead to incorrect rankings of forecasting model performance
\footnote{To be clear, this issue is not specific to \citet{toth2024learning}, but is rather a persistent problem within the time-series forecasting community. Since the default setting in GluonTS has become the de facto standard for benchmarking time-series forecasting models, it has been widely adopted in various studies.}.

\paragraph{Summary.}
The pitfalls are summarised as follows:
\begin{tcolorbox}[colback=blue!10!white]
\begin{enumerate}
    \item[\textbf{Pitfall 1:}] \textbf{Spurious supremacy of quantile-based estimator.} Under the default settings, the quantile-based approach appears superior. However, its convergence behavior is a complex function of both the sample size $M$ and the number of quantile grids $Q$. It is also computationally more expensive than the PWM estimator.
    \item[\textbf{Pitfall 2:}] \textbf{Plug-in bias of PWM estimator}. A naïve MC integration introduces plug-in bias in the CDF term estimation due to the nonlinear nature of the functional.
\end{enumerate}
\end{tcolorbox}
Due to these errors, the current evaluation methods for time-series forecasting models may not accurately reflect their true performance rankings. Therefore, it is necessary to correct the plug-in bias in the PWM estimator to achieve faster and more reliable convergence.

\section{Method: kernel quadrature}
Now, we introduce our approach, an unbiased estimator for the PWM-based CRPS. Any method that can unbias the PWM estimator could be used to address this issue—for example, multi-level Monte Carlo \citep{hong2009estimating, rainforth2018nesting}, but we opted for kernel quadrature based approximation.

\subsection{Unbiased PWM estimator}\label{sec:unbiased}
We introduce the following unbiased estimator:
\begin{align}
    \tilde{h}(y_i, y_j) := \frac{1}{2} \left( y_i \1_{y_i > y_j} + y_j \1_{y_j > y_i} \right).
\end{align}
Note that we use $y_j > y_i$ rather than $y_j \geq y_i$, which naturally ensures zero diagonal elements. As a result, we obtain the unbiased estimator $\mathbb{E}[\tilde{h}(y_i, y_j)] = C(F)$. 

Recall that $y_l$ is the observed value. The simplest way to utilize this unbiased estimator is through Monte Carlo (MC) integration:
\begin{equation}
\begin{aligned}
    \widehat{\text{CRPS}} = 
    &\mathbb{E}_{y \sim \mathbb{P}(f \mid x_i)}[\lvert y - y_l \rvert] + \mathbb{E}_{y \sim \mathbb{P}(f \mid x_i)}[y]\\ 
    &- 2\mathbb{E}_{y, y^\prime \sim \mathbb{P}(f \mid x_i)}[\tilde{h}(y, y^\prime)] \label{eq:u_statistics}
\end{aligned}
\end{equation}

\subsection{Scalable estimator via quantization}
The estimation bias has already been eliminated by the simple solution described above; the remaining challenge is scalability. Although an $\mathcal{O}(M \log M)$ algorithm is asymptotically efficient, it can become impractical for very large $M$ due to memory constraints and increased runtime. To address this, we adopt a kernel quadrature approach that replaces the original set of $M$ equally weighted points with a much smaller collection of $m$ weighted samples such that:
\begin{align}
    \frac{1}{M} \sum_{i=1}^M z(y_i) \approx \sum_{i=1}^m w_i z(y_i), \label{eq:compression}
\end{align}
where $m \ll M$, while ensuring that the approximation error remains minimal. 
Here, we set $z(y)$ as the symmetrized integrand function of Eq.~(\ref{eq:u_statistics})\footnote{
We set $z(y)$ as the following positive and symmetric function ($\xi$ is a constant that ensures the positivity.):
\begin{equation*}
\begin{aligned}
    &k(y_i, y_j \mid y_l) := 
    \tilde{k}(y_i, y_j \mid y_l) + \xi \delta_{y_i, y_j},\\
    \text{where} \quad 
    & \tilde{k}(y_i, y_j \mid y_l) := G(y_i, y_j \mid y_l) - 2 \tilde{h}(y_i, y_j),\\
    &G(y_i, y_j \mid y_l) := \tilde{g}(y_i \mid y_l) + \tilde{g}(y_j \mid y_l),\\
    &\tilde{g}(y_i \mid y_l) := \frac{1}{2}\left(
    \lvert y_i - y_l \rvert + y_i \right),
\end{aligned}
\end{equation*}
}, then the above Eq.~(\ref{eq:compression}) can be understood as compressing the MC integration points into smaller, weighted points (also known as \emph{quantization} \citep{graf2007foundations}). Such a smaller weighted set can exist, given by Tchakaloff’s theorem \citep{tchakaloff1957formules}:
\begin{theorem}[\textbf{Tchakaloff’s theorem}]
    Let $x_1,\cdots,x_m$ be $m$ samples, $w_1,\cdots,w_m \geq 0$ be (positive) weights such that $\sum_{i=1}^{m} w_i = 1$, $\{x_j\}_{j=1}^M = \mu(x)$ be a discrete measure with $M > m$, $\boldsymbol{\varphi} :=(\varphi_1,\cdots,\varphi_n)^\top$ be a $n$-dimensional, integrable, and vector-valued function with $n \leq M + 1$, then there exists a cubature rule
    \begin{align}
        \int_{\mathcal{X}} \boldsymbol{\varphi}(x) \text{d} \mu(x) = \sum_{i=1}^m w_i \boldsymbol{\varphi}(x_i).
    \end{align}\label{thm:cubature}
     such that Eq.~(\ref{thm:cubature}) holds.
\end{theorem}
Notably, this is equality, implying that ``compression'' into smaller weighted points can be performed without introducing approximation errors (lossless compression). The only distinction in our case is that the function $z(y)$ is not vector-valued.

To address this, \citet{hayakawa2022positively} introduced the Nyström method, which approximates the symmetric function as a vector-valued function via eigendecomposition of the Gram matrix, followed by a cubature construction algorithm using recombination. It is based on Carathéodory’s theorem and formulates the problem as one of subset selection: finding the convex hull of random points $(x_j)_{j=1}^m \subset (x_i)_{i=1}^M$. At a high level, the algorithm is conceptually closer to k-means clustering than to classical optimization. The method iteratively:
\begin{compactenum}
    \item Finds a linear dependency among the current support points,
    \item Removes a point while preserving the weighted sum (via pivoting),
    \item Updates the weights accordingly,
    \item Repeats until only  nonzero weights remain.
\end{compactenum}
The computational complexity of this algorithm is\\
$\mathcal{O}(C_\varphi M + m^3 \log (M / m))$, where $C_\varphi$ is the cost of evaluating the functions $(\varphi)_i^n$ at a given point. Thus, it scales linearly with the sample size $M$ and remains computationally efficient. For further methodological details, see~\citet{hayakawa2022positively, adachi2022fast}. 

Since this approach relies on cubature, the only source of error comes from the Nyström approximation. The Nyström method exploits the spectral decay of the Gram matrix, meaning that if the kernel is smooth, the decay is rapid and the error bound remains tight. Empirically, we found that our kernel decays sufficiently fast for large $M$. The Nyström method was originally introduced for kernel quadrature but is more general beyond well-defined Mercer kernels used in typical kernel quadrature, thus we can apply this to the symmetrised matrix by $z(y)$. Yet, this scalable approach remains optional, as using all samples does not introduce additional error. Therefore, it is only necessary when $M$ is too large to handle computationally.

\section{Related work}
\begin{figure*}[t!]
    \centering
    \includegraphics[width=0.85\hsize]{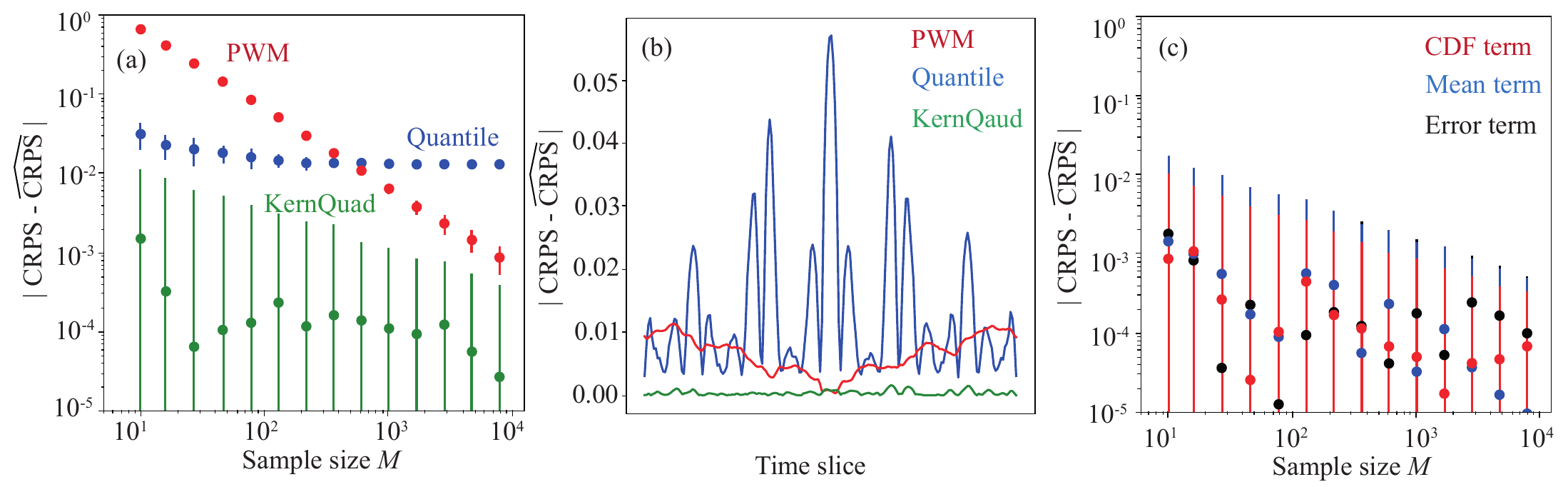}
    \caption{The kernel quadrature (KernQuad) method achieves unbiased estimation. (a) KernQuad outperforms the two widely used alternatives across all sample sizes $M$. (b) KernQuad eliminates bias across time slices. (c) The previously biased CDF term is now at the same level as the other two terms.
    }
    \label{fig:bias}
\end{figure*}
\paragraph{Probabilistic time-series forecasting}
There is a vast array of probabilistic time-series forecasting models. In classical statistical approaches, commonly used models include seasonal naïve, ARIMA, ETS, and linear (ridge) regression~\citep{hyndman2018forecasting}. 
For deep learning models, key representatives for each architecture include: DeepAR~\citep{salinas2020deepar}, based on RNNs, MQ-CNN~\citep{wen2017multi}, based on CNNs, DeepState~\citep{rangapuram2018deep}, based on state-space models,
Transformer-based models~\citep{vaswani2017attention}, leveraging self-attention, and TSDiff~\citep{kollovieh2024predict}, a diffusion-based model, which is considered the current state-of-the-art. 
For GP models, commonly tested methods include: Sparse Variational GP (SVGP) \citep{hensman2013gaussian}, Deep Kernel Learning GPs~\citep{wilson2016deep}, and VRS$^3$GP \citep{toth2024learning}, a recent model using signature kernels, which has demonstrated state-of-the-art performance within the GP framework and achieves comparable accuracy to TSDiff, while requiring significantly shorter training times.

\paragraph{Metric for time-series}
\citet{Candille05, Ferro08, gneiting2007strictly} demonstrated that the CRPS estimator is inherently sensitive to both bias and variance, as CRPS generalizes the absolute error. 
To overcome, various methods have been proposed to mitigate this bias. \citet{Muller05} addressed bias in a CRPS-based skill measure within the specific context of ensemble prediction. \citet{Ferro14} introduced a bias correction factor to improve the fairness of CRPS for ensemble forecasts, accounting for finite ensemble sizes. \citet{ZamoNaveau18} further reviewed CRPS estimators derived from limited sample information, providing practical guidelines for selecting the optimal estimator based on the type of random variable. 
Unlike these post-processing corrections aimed at reducing bias in CRPS estimators, our study constructed an unbiased CRPS estimator.

\paragraph{Kernel quadrature}
There are several kernel quadrature algorithms, including herding/optimization \citep{chen2010super, huszar2012optimally}, random sampling \citep{bach2017on}, determinantal point processes (DPP; \citet{belhadji2019kernel}, recombination~\citep{hayakawa2022positively}. While any of these methods can be applied to our problem, their primary focus is on selecting quadrature nodes, rather than debiasing. 

\section{Results}
\subsection{Experimental setup}
We implemented our code using PyTorch~\citep{paszke2019pytorch} and GPyTorch~\citep{gardner2018gpytorch} for modeling Gaussian processes (GPs). The implementation of scalable kernel quadrature is based on SOBER~\citep{adachi2024quadrature}, but our method is not limited to this library. 
All experiments were averaged over 100 repeats for the Ackley function and 3 repeats for the multi-sinusoidal wave datasets, each with different random seeds. The experiments were conducted on a MacBook Pro (2019), 2.4 GHz 8-Core Intel Core i9, 64 GB.

\subsection{Unbiased estimator}
We confirmed that the bias issues identified in Section~\ref{sec:pitfall} have been resolved using our kernel quadrature (KernQuad) approach. Fig.~\ref{fig:bias} clearly demonstrates that KernQuad achieves unbiased estimation. 
The previously observed bias over time slices has disappeared, leaving only location-independent noise. Additionally, the CDF term, which was previously the primary bottleneck, is no longer a limiting factor, as it now exhibits the same convergence rate as the other error terms. 
These results clearly indicate that KernQuad is unbiased and consistently outperforms the two existing baselines.

\subsection{Time-series forecasting models}
\begin{figure*}[t!]
    \centering
    \includegraphics[width=0.75\hsize]{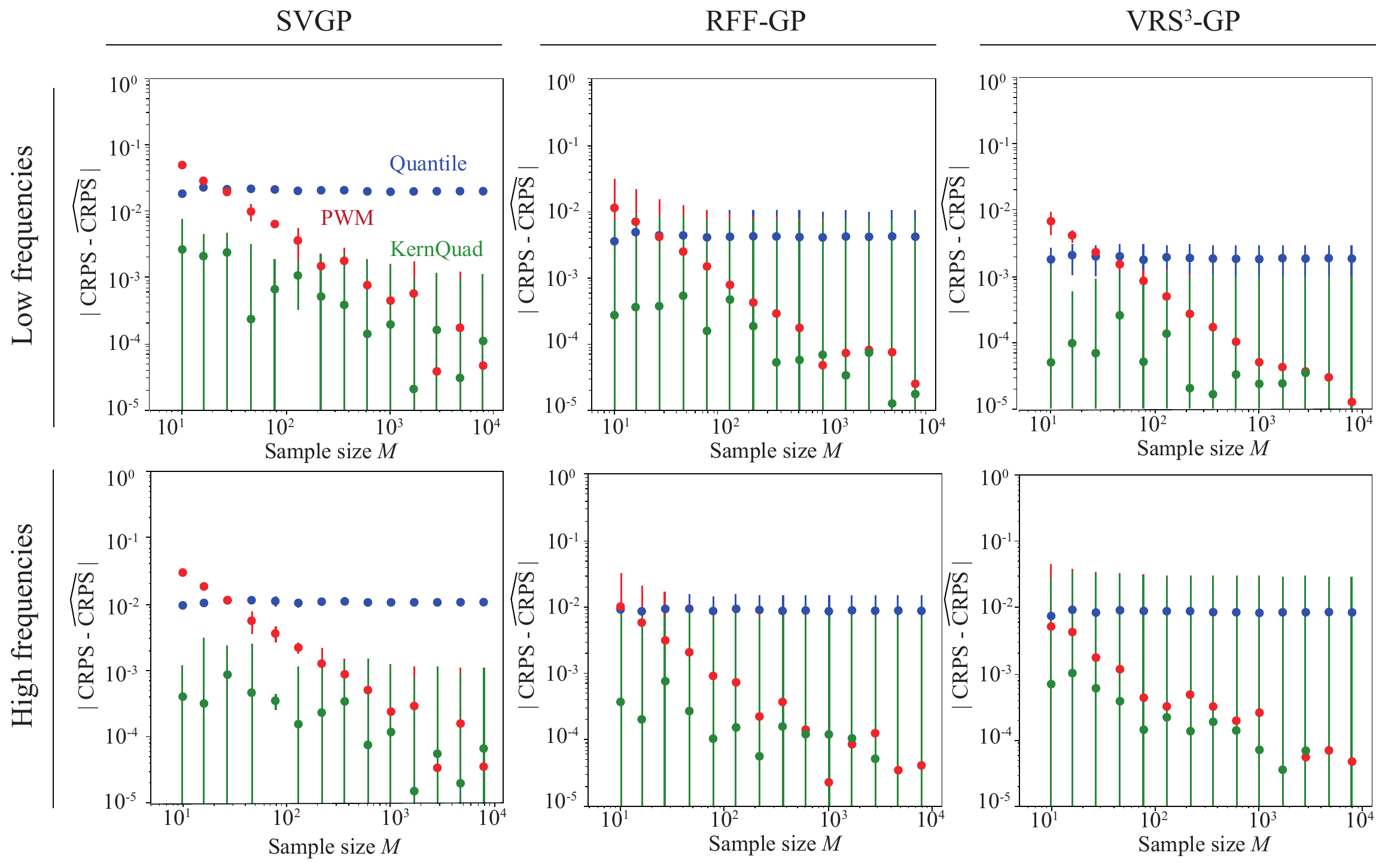}
    \caption{The kernel quadrature method consistently outperforms the quantile-based and PWM estimators across three forecasting models and two test datasets.
    }
    \label{fig:model_comp}
\end{figure*}
\begin{table}
    \centering
    \caption{Comparison of four CRPS estimators on the high-frequency multi-sinusoidal test dataset across three forecasting models. The quantile estimator incorrectly ranks SRV$^3$ as performing worse than RFF-GP on this dataset. The actual values should be scaled by $\times 10^{-3}$.}
    \resizebox{0.5\textwidth}{!}{
    \begin{tabular}{lcccc}
        \toprule
         method &  SVGP & RFFGP & VRS$^3$GP\\
         \midrule
         closed-form & 8.1401 $\pm$ 0.9767 & 2.424 $\pm$ 0.0081 & 2.419 $\pm$ 0.0094\\
         quantile & 9.0332 $\pm$ 1.1105 & $2.701 \pm 0.0240$ & 2.702 $\pm$ 0.0254 \\
         PWM & 8.2144 $\pm$ 1.0066 & $2.409 \pm 0.0326$ & 2.397 $\pm$ 0.0468\\
         KernQuad & 8.1110 $\pm$ 0.9264 & $2.424 \pm 0.0147$ &  2.422 $\pm$ 0.0090 \\
         \bottomrule
    \end{tabular}
    }
    \label{tab:model_comp}
\end{table}
We further evaluate the performance of KernQuad across various time-series forecasting models. Among the available models, we select SVGP~\citep{hensman2013gaussian}, random Fourier feature GP (RFF-GP), and variational recurrent sparse spectrum signature GP (VRS$^3$GP; \cite{toth2024learning}) for comparison. The primary reason for choosing these models is that GP-based methods allow for the computation of true CRPS, enabling us to directly assess estimation errors. Notably, VRS$^3$GP has demonstrated state-of-the-art performance in time-series forecasting tasks, as shown in Table 1, making this a practically relevant setting.

To analyze performance dependencies, we test on synthetic time-series data generated from multi-sinusoidal waves with four weighted components of different frequencies but no phase shift. We prepare two synthetic test functions: (a) Low-frequency waves $[0.1,1,2,5]$, and (b) High-frequency waves $[1,5,10,20]$ for the $L=800$ training and $T=100$ test time steps.  
By definition, learning high-frequency components is easier than low-frequency ones, as the latter appear less frequently. Thus, the low-frequency dataset is more challenging, making it easier to distinguish model performance differences. Conversely, the high-frequency dataset is easier for most models, making it harder to differentiate model performance, and thus CRPS estimation accuracy becomes more critical. To ensure robustness, we repeat model training three times with different random seeds. Consequently, even the closed-form CRPS estimator is computed as a MC integration over three samples. 

Fig.~\ref{fig:model_comp} presents the convergence rates across three forecasting models and two datasets. The trends remain consistent across all datasets and models: Our kernel quadrature method consistently outperforms the two other CRPS estimators. The quantile-based estimator, which is the current default approach, performs the worst. The estimation error from the quantile estimator leads to incorrect model rankings. Table~\ref{tab:model_comp} further highlights this issue, showing that the quantile estimator erroneously ranks RFF-GP as outperforming VRS$^3$GP. While the difference between the two models falls within the standard deviation, making this specific instance detectable by closely examining error bars, it serves as a clear counterexample where the quantile-based CRPS estimator can misrank models. This further motivates the use of our kernel quadrature estimator instead.

Although this issue is identifiable in GP-based forecasting models, where we can numerically verify CRPS approximation errors, the same verification is not possible for deep learning models. As seen in Table 1, the standard deviations of deep learning models are typically larger than those of GP-based methods, and their variability is unpredictable, as different methods exhibit the largest standard deviation on different datasets. 
Thus, we argue that fixing this issue with our kernel quadrature approach is essential for ensuring more reliable model performance comparisons.

\section{Conclusion and Limitations}
We first identify the pitfalls of the two existing CRPS estimators; the quantile-based and PWM estimators. The quantile estimator, which is the current default in GluonTS \citep{alexandrov2020gluonts}, exhibits a consistent bias that cannot be eliminated by increasing the sample size. The PWM estimator suffers from plug-in bias, which weakly converges to the true CRPS in the infinite sample limit. However, this bias remains the bottleneck in its estimation process. 
To address the pitfalls, we propose an unbiased estimator and its scalable approximation using kernel quadrature. Our proposed unbiased estimator consistently achieves lower estimation errors across all sample sizes, three datasets, and three forecasting models. Moreover, we demonstrated that the quantile estimator can lead to incorrect model rankings on certain datasets, whereas our kernel quadrature estimator preserves the correct rankings. This highlights the importance of minimizing approximation errors in time-series forecasting model evaluation.

Similar CDF-based estimators exist for other probabilistic metrics, such as the energy score \citep{fahy1994statistical}, calibration score \citep{NEURIPS2024_9961e426, fujisawa2025pacbayes}, conformal prediction \citep{snell2025conformal}, and spectral risk measure \citep{pandey2021estimation}. We anticipate that our approach can be extended to these metrics, enabling more reliable and unbiased estimation in broader applications. Recent techniques from probabilistic numerics, such as those proposed by \citet{wenger2020non, adachi2024adaptive}, also present promising directions for further extension.

Although kernel quadrature offers scalable computation, it still introduces additional approximation error. Also, the inherent convergence rate is limited by the standard MC rate $\mathcal{O}(1/\sqrt{M})$. Leveraging the faster convergence rate of standard Bayesian quadrature (BQ) is a promising direction, yet it produces the following challenge: 
\begin{compactenum}
    \item Applying BQ to the error and mean terms is straightforward, but the CDF term is challenging. The CDF is monotonic and bounded, whereas GPs do not inherently impose such constraints. Although prior work addresses these limitations, the methods are often computationally expensive. In time-series applications, where datasets can contain millions of time points, CRPS estimation must be repeated millions of times per random seed. Hence, computational efficiency is critical.
    \item Constructing suitable $(x,y)$ pairs is non-trivial. Here, $x$ represents samples like $f(x_l)$, but obtaining the corresponding ``ground-truth CDF'' values for $y$ is difficult. We tried using empirical CDF values as $y$ for GP training and applied BQ, but the results were worse than MC integration. If one uses a standard kernel like RBF, the task ends up being more similar to kernel density estimation than BQ. Recent paper, \citet{snell2025conformal}, which takes a promising alternative approach by placing a Dirichlet prior on quantile spacing. While it still requires MC integration, it may be worth exploring in future work.
\end{compactenum}

\section*{Acknowledgements}
We thank anonymous reviewers for their valuable feedback and discussions. MA is supported by the Clarendon Fund, the Oxford Kobe Scholarship, the Watanabe Foundation, and Toyota Motor Corporation. MF is supported by KAKENHI (Grant Number: 25K21286). 

\bibliography{references}

\end{document}